\pdfoutput=1
\documentclass[11pt,UTF8]{article} 

\usepackage{booktabs}
\usepackage{cuted}
\usepackage{balance}
\usepackage{amsmath,lipsum, amssymb} %
\usepackage{multirow}
\usepackage{multicol}
\usepackage{bbding}
\usepackage{resizegather}
\usepackage[linesnumbered,ruled,vlined]{algorithm2e}
\usepackage{float}
\usepackage[final]{acl} %

\usepackage{times}
\usepackage{latexsym}
\usepackage{CJKutf8}

\usepackage[T1]{fontenc} 

\usepackage[utf8]{inputenc}

\usepackage{microtype}

\usepackage{inconsolata}

\usepackage{graphicx}

\title{NALA: an Effective and Interpretable Entity Alignment Method}

\author{Chuanhao Xu, Jingwei Cheng\thanks{Corresponding author}, Fu Zhang \\ 
  School of Computer Science and Engineering, Northeastern University, China \\
  2201892@stu.neu.edu.cn, \{chengjingwei, zhangfu\}@mail.neu.edu.cn
  }

\begin{document}
\begin{CJK*}{UTF8}{gbsn}
\maketitle
\begin{abstract}
Entity alignment (EA) aims to find equivalent entities between two Knowledge Graphs. 
Existing embedding-based EA methods usually encode entities as embeddings, triples as embeddings' constraint and learn to align the embeddings. 
However, the details of the underlying logical inference steps among the alignment process are usually omitted, 
resulting in inadequate inference process. 
In this paper, we introduce NALA, an entity alignment method that captures three types of logical inference paths with Non-Axiomatic Logic (NAL). 
Type \uppercase\expandafter{\romannumeral1}\&\uppercase\expandafter{\romannumeral2} align the entity pairs 
and type \uppercase\expandafter{\romannumeral3} aligns relations.
NALA iteratively aligns entities and relations by integrating        
the conclusions of the inference paths. 
Our method is logically interpretable and extensible by introducing NAL, and thus suitable for various EA settings.
Experimental results show that NALA outperforms state-of-the-art methods in terms of Hits@1, achieving 0.98+ on all three datasets of DBP15K with both supervised and unsupervised settings. 
We offer a pioneering in-depth analysis of the fundamental principles of entity alignment, approaching the subject from a unified and logical perspective.
Our code is available at \href{https://github.com/13998151318/NALA}{https://github.com/13998151318/NALA}.
\end{abstract}

\maketitle

\section{Introduction}
\begin{figure}
  \centering
  \includegraphics[width=170pt]{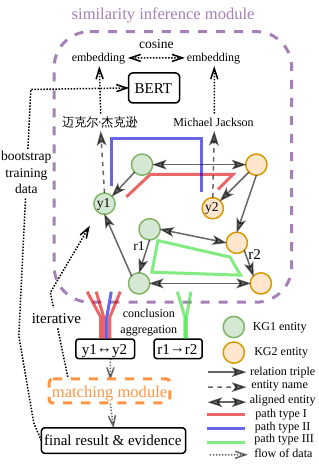} 
  \caption{An overview illustration of NALA.}
  \label{fig:overview}
\end{figure}
Knowledge graphs (KGs), which store massive facts about the real world,
expresses massive information in a form closer to human cognition.
KGs can be used by various application domains, such as question answering, recommender systems and language representation learning 
(knowledge graph enhanced language model) ~\cite{RN235, RN250}. 
The information contained in each individual KG project, such as DBpedia ~\cite{RN237} and YAGO ~\cite{RN238} is limited. 
So the task of entity alignment (EA) is proposed to increase KG completeness. 
The EA task consists of integrating two or more KGs into a same KG by aligning nodes that refer to the same entity.

There are many embedding-based EA methods ~\cite{RN221} that leverage deep learning techniques to represent entities with low-dimension\-al
embeddings, and align entities with a similarity function on the embedding space. KGs' triples and seed alignments are usually seen as embeddings' 
constraint during the training process of such embedding model.
The structural and side information of KGs are usually utilized via embedding propagation, aggregation or interaction.
Generally speaking, there are some crucial shortcomings of embedding-based EA methods: 
\textit{First}, they possibly lack complex reasoning capability. 
Some of them are enhanced by paths ~\cite{RN240}, however, due to the nature of vector representation, it is not easy to perform or approximate symbolic reasoning on such paths. 
\textit{Second}, they lack interpretability in the models, so they have to rely solely on numerical evaluation metrics to evaluate their performance.
Thus the cons and pros of their model design may not be properly evaluated.
\textit{Third}, the absence of a unified framework explaining the mechanism of embedding learning and processing renders their semantic or structural learning capability quite mysterious.

Apart from embedding-based methods, path-based methods directly estimates entity similarities from the contextual data (path) that
are available in the two input KGs. 
A "path" usually refers to an interconnected sequence of edges that links two entities of different KGs. 
The edges can be either relations or entity similarities. 
We refer to the estimation of entity similarities by processing and aggregating the paths as "similarity inference".
There is a potential advantage that path-based methods can capture fine-grained matches of neighbors while the traditional embe\-dding-based methods can't.
There are also emerging methods that combine the idea of embedding learning and path reasoning.
More recently, path-based (such as PAR\-IS+ ~\cite{RN146}) and combined methods (such as BERT-INT ~\cite{RN168} and FGWEA ~\cite{RN163}) 
are starting to surpass the performance of traditional em\-bedding-based methods. 
However, they failed to handle the similarity inference appropriately to some extent, 
possibly due to the lack of proper formalization of the inference paths and steps.

To address the aforementioned issues of existing methods, we carefully 
examine the similarity inference of EA from the logical perspective.
Thus we propose a path-based EA method NALA, 
where NAL stands for Non-Axiomatic Logic ~\cite{RN219} and "A" for align. 
NAL is a term logic with a specific semantic theory and its design suits KG tasks (see Section \ref{Why NAL}).

As illustrated in \autoref{fig:overview}, NALA adopts an iterative entity alignment strategy with two modules, namely \textbf{similarity inference module} and \textbf{matching module}.
For each iteration, it first performs the similarity inference module.
The module has three functions: 1) Search for three types of paths across the KGs. 2) Inference on the path instances with fixed inference rules defined by NAL. 3) Aggregate the conclusions of the paths. 
\textit{Type \uppercase\expandafter{\romannumeral1}\&\uppercase\expandafter{\romannumeral2} paths} yield conclusions on similarities of entity pairs. 
\textit{Type \uppercase\expandafter{\romannumeral3} paths} yield conclusions on substitutability (or, inheritance) of relation or attribute pairs.
We use BERT embedding to assist path inference by obtaining similarity among entity names and attribute values, which constitutes some premises of the paths.
Then NALA uses the matching module to obtain 1-to-1 EA results of the current iteration.
We propose an algorithm, namely rBMat algorithm with swapping step (see Section \ref{Matching Module}) for the matching module.

Experiments on cross-lingual EA dataset DBP15K demonstrate that NALA outperforms SOTA EA
methods in 5 different setting groups (including both supervised and unsupervised scenarios),
 showcasing the effectiveness of our proposed logical similarity inference module and matching module. 
Ablation study shows that our design choices jointly boost the overall performance of NALA.


Our contributions can be summarized as:
\begin{itemize}
\item 
We propose an interpretable EA framework NALA, which tackle the EA problem with similarity inference phase and matching phase. 
Various types of logical paths are formalized within the similarity inference phase. 
\item
NALA aligns entities and relations simultaneously with a unified yet extensible logical framework. 
\item 
Our framework bridges the gap between embedding-based and path-based EA. 
\item 
Our proposed method achieves SOTA on a widely used EA dataset DBP15K's various settings.
\item 
We present the first in-depth analysis of EA's basic principles from a unified logical perspective, 
and help explain the mechanism of other EA methods.

\end{itemize}

\section{Preliminaries} \label{Preliminaries}
\subsection{Knowledge Graph and Entity Alignment}
\textbf{KGs}. Knowledge graphs (KGs) are knowledge bases that store knowledge in the form of triples (or "facts").
We refer to (head, relation, tail) and (head, attribute, literal) as relation and attribute
triples, respectively. Examples of both triple types are (New\_\-Zeala\-nd,
capital, Wellington) and (New\_\-Zealand, establishedDate, "1947-11-25"), respectively. 
To summarize, a KG is characterized with a number of relation triples from $ \mathcal{E} \times \mathcal{R}  \times \mathcal{E}$ 
and a number of attribute triples from $ \mathcal{E} \times \mathcal{A} \times \mathcal{L}$ 
, where $ \mathcal{E}, \mathcal{R}, \mathcal{A}$, and $\mathcal{L}$
indicate the set of entities, relations, attributes and literals, respectively.

\textbf{EA}. The entity alignment (EA) problem is typically defined between two
KGs, $\mathcal{KG}_1$ and $ \mathcal{KG}_2$,
where the task consists of finding equivalences (so-called alignment)
between the set of entities $\mathcal{E}_1$ and $\mathcal{E}_2$ of the two KGs. Sometimes
there exists a set of given equivalences that can be used as supervision. This set $\mathcal{S}$ is known as seed alignment set.
We assume that there exists a ground truth set $\mathcal{G} = \{(x_1, x_2) \in  \mathcal{E}_1 \times \mathcal{E}_2 \vert\ x_1 \equiv x_2\}$  that
includes all known equivalences between pairs of entities.
We use the ground truth set to evaluate the performance of our method.
We use the subscript of an entity identifier ($x$ or $y$) or a relation identifier ($r$) to represent which KG it comes from.


\subsection{Represent KGs with NAL} \label{Represent KGs with NAL}

A brief introduction to NAL is presented in Appendix \ref{Introduction of NAL}.

In this paper, every entity, literal or relation is regarded as an \textit{atomic term} in NAL. 
Triple ($x$,\ $r$,\ $y$) is  
reinterpreted as \textit{inheritance statement} (*, $x$,\ $y$) $\rightarrow$\ $r$. Its intuitive meaning is "The relation between $x$ and $y$ is a specialization of relational term $r$". 
The triples (or "facts") of the KGs can be seen as absolutely true (for \textit{frequency}) and with sufficient evidence (for \textit{confidence}) to some extent, so the \textit{truth-value} attached to the 
\textit{statement} is $\left \langle 1, 1  \right \rangle$. Entity equivalency $\ x_1 \equiv x_2$ can be seen as an extreme case of entity similarity $x_1$\ $\leftrightarrow$\  $x_2\ $, 
so we align entities by similarity inference.
As for relations, the \textit{inheritance statement} $r_1$\ $\rightarrow$\ $r_2$ intuitively represents a correspondence of two relations of different KGs
such that one relational fact of $r_1$ in $\mathcal{KG}_1$ implies the existence of a corresponding relational fact of $r_2$ in $\mathcal{KG}_2$.

We automatically duplicates every original KG triple ($x$,\ $r$,\ $y$) with a reversed triple ($y$,\ $r^{-1}$,\ $x$) upon KG loading, where $r^{-1}$ represents the reverse relation or attribute of $r$. 

\textbf{Inference path}. We define an instance of inference path as a premise set of NAL \textit{sentences} (triples, similarities, etc.)
and a series of corresponding inference steps which will eventually lead to a conclusion \textit{sentence}.
The premise \textit{sentences} are either in the KGs or inferred from the KGs.
A type of inference path is a shared form of paths and it can be instantiated with concrete entities and relations. 
It is usually utilized for a certain purpose, such as aligning entities or aligning relations.

\subsection{Why NAL} \label{Why NAL}
Actually there might be many different logical systems that are qualified to represent the similarity inference process of EA.
However, we believe that the non-axiomatic nature of NAL fits in the domain of knowledge graph better than those axiomatic logical systems, because real world KGs need to deal with the problem of open-domain
 and alterable, incomplete or conflicting facts. Fundamentally, the tasks of knowledge graph (such as EA), fits well with the assumption
of insufficient knowledge and resources ~\cite{RN219}, which is the basic assumption of NAL.

Technically speaking, NAL can represent entities, relations and relational triples, which are essential for EA. It can also perform formal reasoning and evidence aggregation, which is useful to align entities.
The \textit{frequency}/\textit{confidence} measurement of \textit{truth-value} is suitable to represent fuzziness and unknownness in the similarity inference process. The high expressiveness of NAL makes our approach extensible, which may benefit subsequent studies.

\subsection{Related Work of EA} \label{Related Work}

Generally speaking, there are three families of EA methods: embe\-dding-based, path-based and combined methods, as elaborated in this section.

In recent years, embedding-based methods have become mainstre\-am for addressing the EA task ~\cite{RN163, RN221}.
Their main idea is to embed the nodes (entities) and edges (relations or attributes) of a KG into a
low-dimensional vector space that preserves their similarities in the original KG.
Embedding-based methods may suffer from the negative influence from the dissimilar neighbors, according to ~\cite{RN168}.

In addition to embedding-based methods, there exist path-based methods that directly estimates entity similarities from the contextual data (path) that
are available in the two input KGs. 
The distinction between embedding-based and path-based methods is sometimes obscure.

There are also emerging methods that combine the idea of embedding learning and path reasoning.
More recently, path-based and combined methods are starting to surpass the performance of traditional em\-bedding-based methods.

Our proposed method NALA inherits and develops the ideas of two path-based methods PARIS  ~\cite{RN216} and PARIS+. The two methods 
as well as some other EA methods that will be compared with our results are introduced in Appendix \ref{Related Works appendix}.

\section{The Proposed Method} \label{The Proposed Method}

The overall structure of NALA (as illustrated in \autoref{fig:overview} and Algorithm \ref{NALA(supervised)}) adopts an iterative aligning strategy, and for each iteration
it first performs similarity inference, 
then it uses the matching module (rBMat algorithm with swapping step in Section \ref{Matching Module}) to obtain EA results.
The inference within each iteration benefits from the alignment results (both entities and relations) of the previous iteration. 

\begin{figure*}[htbp]
\begin{gather*}
\frac{\left(*,x_1,y_1\right) \rightarrow r_1  \hspace{5mm} (1) ,\ r_1 \rightarrow r_2  \hspace{5mm} (2)}
{\left(*,x_1,y_1\right) \rightarrow r_2  \hspace{5mm} (3)}Deduction \\[1.5ex]
\frac{\left(*,x_2,y_2\right) \rightarrow r_2 \hspace{5mm} (4) ,\ x_1 \leftrightarrow x_2  \hspace{5mm} (5)}
{\left(*,x_1,y_2\right) \rightarrow r_2  \hspace{5mm} (6)}Analogy* \\[1.5ex]
\frac{( \left(*,\#a,\$b\right) \rightarrow \#r\ \wedge\ \left(*,\#a,\$c\right) \rightarrow \#r\ \nonumber 
    \wedge\ \#r \rightarrow [fun] ) \ \Rightarrow \$b \leftrightarrow \$c \hspace{5mm} (7) ,\ \left(*,x_1,y_1\right) \rightarrow r_2  \hspace{5mm} (3)}
{( \left(*,x_1,\$c\right) \rightarrow r_2\ \nonumber 
    \wedge\ r_2 \rightarrow [fun] ) \ \Rightarrow y_1 \leftrightarrow \$c \hspace{5mm} (8)}Conditional\ deduction \\[1.5ex]
\frac{( \left(*,x_1,\$c\right) \rightarrow r_2\ \nonumber 
    \wedge\ r_2 \rightarrow [fun] ) \ \Rightarrow y_1 \leftrightarrow \$c \hspace{5mm} (8) ,\ \left(*,x_1,y_2\right) \rightarrow r_2  \hspace{5mm} (6)}
{ r_2 \rightarrow [fun] \ \Rightarrow y_1 \leftrightarrow y_2 \hspace{5mm} (9)}Conditional\ deduction \\[1.5ex]
\frac{r_2 \rightarrow [fun] \ \Rightarrow y_1 \leftrightarrow y_2 \hspace{5mm} (9) ,\ r_2 \rightarrow [fun]  \hspace{5mm} (10)}
{y_1 \leftrightarrow y_2  \hspace{5mm} (11)}Conditional\ deduction
\end{gather*}
\caption{NAL formalization of \textit{type \uppercase\expandafter{\romannumeral1}\ path}} \label{NAL formalization of type 1 path}
\end{figure*}

\subsection{Similarity Inference Module} \label{Similarity Inference Module}
We formalize the similarity inference module as using 
NAL's \textit{revision} inference rule to aggregate three types of inference paths.
The first two types calculates similarity for entities and the third type for relations.
We register evidential information while performing path inference, that is memorizing which premises constitute the 
specific path instance and such information will be used to generate evidence log file.

\subsubsection{\textit{Type \uppercase\expandafter{\romannumeral1} Path}: Align Entities by Triples} \label{type 1 path: Align Entities by Triples}

Inspired by the probabilistic alignment method of PARIS, we formalize the key point of the similarity inference process as \textit{type \uppercase\expandafter{\romannumeral1} path}. 
\textit{Type \uppercase\expandafter{\romannumeral1} paths} are bridge-like inference paths between to-be-aligned entity pairs. 
Valid \textit{type \uppercase\expandafter{\romannumeral1} paths} are retrieved from the KGs in a depth-first manner.
The NAL formalization is represented in a form similar to natural deduction \cite{RN275}, as shown in \autoref{NAL formalization of type 1 path}. Each step of inference is characterized by 
two premises (on the top of the inference line) and a conclusion (on the bottom of the inference line).
The inference rule is indicated on the right edge of the inference line.

First, we elaborate the premises (1, 2, 4, 5, 7 and 10). Premise (1) and (4) can simultaneously be relational triples, or attribute triples, where $x_1$ and $x_2$ are either entities or literals respectively. 
As for premise (5), in the case of entity pair, the \textit{similarity statement} comes from either seed alignments or alignments of the previous iteration. 
We omit any entity \textit{similarity statement} which has a $f$ or $c$ lesser than $theta$, a hyper-parameter.
And in the case of literal pair, see Section \ref{type 2 path: Align Entities by Name}.  
The relation inheritance of premise (2) is inferred in Section \ref{type 3 path: Aligning Relations}. 
PARIS evaluates the degree of functionality of relation $r_2$ with precomputed functionalities of each relation. 
We interpret it as an \textit{inheritance statement} $r_2$\ $\rightarrow$\ [$fun$] with the degree reflected in the \textit{truth-value}, that is, premise (10). 
The statement intuitively means "$r_2$ has the functional property (to some extent)".
Premise (7) is an implication statement that is regarded as a definition or a piece of essence of the concept "functionality". 
The functionalities of relations seems to reflect a widespread orderliness of reality or human cognition and we leverage such orderliness.

With the premises and their known \textit{truth-values}, we performs \textit{syllogistic} inference according to the fixed rule table \autoref{tab:truth functions}.
Statement (11) is the conclusion of the inference steps and the steps act as a summarizing or validation process of the premises.

\begin{figure}
  \centering
  \includegraphics[width=210pt]{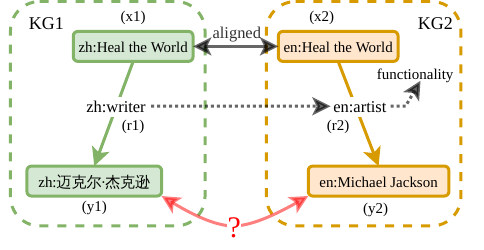} 
  \caption{An instance of \textit{type \uppercase\expandafter{\romannumeral1} path}, fetching from DBP15K zh-en and omitting irrelevant triples. Grey dashed arrow represents inheritance 
  between the relations and "functionality".}
  \label{fig:path1}
\end{figure}

The idea of \textit{type \uppercase\expandafter{\romannumeral1} path} can be explained with an example as shown in \autoref{fig:path1}. We would like to figure out whether "zh:迈克尔·杰克逊" and "en:Michael Jackson" 
refers to the same entity. We find out that a related pair of entity: "zh:Heal the World" and "en:Heal the World" are known aligned entity pair (or, inferred to be aligned). "zh:Heal the World"'s writer is "zh:迈克尔·杰克逊" and "en:Heal the World"'s artist is "en:Michael Jackson". 
We also know that being the writer of something probably implies being the artist of it (the relation inheritance). 
We have looked through the KG and found out that a certain work usually has only one artist. We conclude that these premises together form a certain amount of positive evidence that supports "zh:迈克尔·杰克逊" and "en:Michael Jackson" being the same entity.
\textit{Type \uppercase\expandafter{\romannumeral1} path} can be seen as the fundamental entity alignment evidence (signal).


The conclusions with the same statement but obtained from different \textit{type \uppercase\expandafter{\romannumeral1} paths} are merged by \textit{probabilistic revision} rule because of the probabilistic nature of functionality. For example, the functionality of relation "zh:writer" is 0.78 which means that the majority of works approximately have one to two writers. While reasoning with \textit{type \uppercase\expandafter{\romannumeral1} paths}, we could not know how many writers does "zh:Heal the World" have, the conclusion has a probabilistic nature because we don't know whether "zh:迈克尔·杰克逊" and "en:Michael Jackson" is the same writer of "Heal the World". 
The \textit{probabilistic revision} rule is similar with the 
continued multiplication of PARIS's formula for $Pr\left( x_1 \equiv x_2 \right)$ (given in Appendix \ref{Related Works appendix}), 
except for the introduction of \textit{confidence}.
We have some additional remarks of the path in Appendix \ref{Additional Remarks on Inference Paths}.

\subsubsection{\textit{Type \uppercase\expandafter{\romannumeral2} Path}: Align Entities by Name} \label{type 2 path: Align Entities by Name}
\textit{Type \uppercase\expandafter{\romannumeral2} path} is the direct path linking the to-be-aligned entities with their name/desc\-ription similarity.
It only has a conclusion statement:

\centerline{$y_1 \leftrightarrow y_2 \ \  \left \langle sim(name(y_1), name(y_2)),\ C_{name}\right \rangle$} 

\noindent where $sim$ is the cosine similarity of entity name/description embedding and $C_{name}$ is a hyper-parameter. 
NALA adopts BERT as the embedding model. The BERT unit is finetuned on the name/description of seed alignment entity pairs before embedding generation, similar with BERT-INT.
The conclusion of a \textit{type \uppercase\expandafter{\romannumeral2} path} is seen as a piece of evidence and fused with other evidences by \textit{revision} rule.

We implement an adaptive method to automatically set $C_{name}$ to avoid excessive parameter tuning. First, 
with a specific setting and dataset, we run NALA for 5 iterations (with a default $C_{name} = 0.5$ which represents a unit amount of evidence) and calculate the alignment output's average \textit{confidence}.
We set $C_{name} = halve\_evidence(average\_confidence)$, where $halve\_evidence$ is a function that outputs a \textit{confidence} value 
that corresponds to half of evidence amount of the input \textit{confidence}. The idea is to balance the influence of structural 
information and name information, preventing the name information's evidence from being too strong or too weak. Then we restart NALA from the first iteration and $C_{name}$ remains unchanged.
If translated name is available, the evidence amount is equally divided between translated and original name's $C_{name}$.
If the BERT unit is un-finetuned, we penalize $C_{name}$'s evidence amount by a factor $C_{penalty}$.

We also obtain attribute value embedding with the BERT unit and their cosine similarities are used to convert to the \textit{truth-value} of premise (5) where $x_1$ and $x_2$ are distinct attribute values: 

\centerline{$x_1 \leftrightarrow x_2\ \left \langle f = sim(x_1, x_2),\ c = \-sim(x_1, x_2)\right \rangle$}
\noindent The idea is that the pair of similar embedding of the deep learning model which has higher similarity is usually more verifiable. 
For identical attribute values, the \textit{truth-value} is simply $\left \langle 1,\ 1\right \rangle$.
There are thousands of distinct attribute values in a KG, so for an attribute value we only consider the $K_{value}$ most similar (but not identical) values in the other KG to prevent an explosive number of value similarities.
$K_{value}$ is a hyper-parameter and in implementation we set $K_{value}$ to 1. 
See more discussion of utilizing literal value and \textit{type \uppercase\expandafter{\romannumeral2} path} in appendix \ref{Discussion Literal value} and \ref{Discussion type 2 path}.

\subsubsection{\textit{Type \uppercase\expandafter{\romannumeral3} Path}: Aligning Relations} \label{type 3 path: Aligning Relations}
NALA align relations by path inference, which is a different approach from PARIS's probabilistic relation aligning method.
We formalize the inference process as \textit{type \uppercase\expandafter{\romannumeral3} path} as shown in \autoref{NAL formalization of type 3 path}.
The premises are (12, 13 15 and 17) and the conclusion is (18).

\begin{figure}[htbp]
\begin{gather*}
\frac{\left(*,x_1,y_1\right) \rightarrow r_1  \hspace{5mm} (12) ,\ x_1 \leftrightarrow x_2  \hspace{5mm} (13)}
{\left(*,x_2,y_1\right) \rightarrow r_1  \hspace{5mm} (14)}Analogy \\[1.5ex]
\frac{\left(*,x_2,y_1\right) \rightarrow r_1  \hspace{5mm} (14) ,\ y_1 \leftrightarrow y_2  \hspace{5mm} (15)}
{\left(*,x_2,y_2\right) \rightarrow r_1  \hspace{5mm} (16)}Analogy \\[1.5ex]
\frac{\left(*,x_2,y_2\right) \rightarrow r_2  \hspace{5mm} (17) ,\ \left(*,x_2,y_2\right) \rightarrow r_1  \hspace{5mm} (16)}
{r_1 \rightarrow r_2  \hspace{5mm} (18)}Induction 
\end{gather*}
\caption{NAL formalization of \textit{type \uppercase\expandafter{\romannumeral3} path}} \label{NAL formalization of type 3 path}
\end{figure}

There are two versions of \textit{type \uppercase\expandafter{\romannumeral3} path} and the only difference is the 
\textit{truth-value} of premise (17). The positive version's \textit{truth-value} is 
$\left \langle 1,\ 1\right \rangle$ and the negative version's is $\left \langle 0,\ C_{absent}\right \rangle$, where $C_{absent}$ is 
a hyper-parameter for absent or missing fact. We argue that when there is a fact present in the KG, it is usually confident. However,
when there is an absent fact in the KG, its denial is not as confident because the KG may be incomplete. In implementation we set $C_{absent} = 0.5$ 
(which represents a unit amount of evidence).

The \textit{induction} inference rule in \textit{type \uppercase\expandafter{\romannumeral3} path} is a \textit{weak inference rule}, so the upper bound of its conclusion's \textit{confidence} is lower than the 
\textit{strong inference rules} (such as \textit{deduction} and \textit{analogy}).
The positive version only generates positive evidence for the conclusion and the negative version only generates negative evidence, 
because of the characteristic of the \textit{induction} rule.

\begin{figure}
  \centering
  \includegraphics[width=210pt]{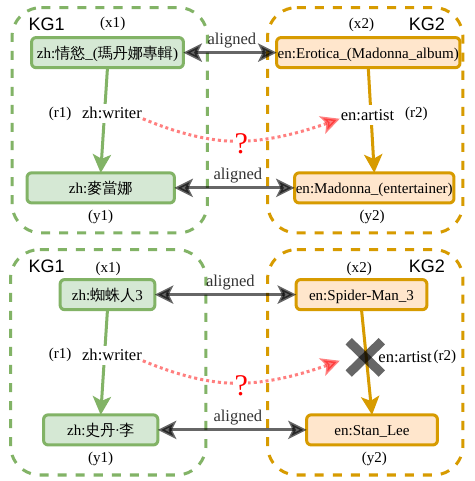} 
  \caption{An illustration of \textit{type \uppercase\expandafter{\romannumeral3} path}. The upper half represents positive version of the path
  and the lower half represents negative version. The dark cross represents the absence of the triple.}
  \label{fig:path3}
\end{figure}

Two instances of \textit{type \uppercase\expandafter{\romannumeral3} path} are illustrated in \autoref{fig:path3}. 
We would like to figure out the inheritance between relations "zh:writer" and "en:artist", so multiple path instances are collected, 
including both positive version and negative version of the type of path. 
The conclusions of the two versions are supposed to be merged by the \textit{revision} rule.
The relation inheritance sentence of \textit{type \uppercase\expandafter{\romannumeral1} path} use computation result of 
\textit{type \uppercase\expandafter{\romannumeral3} path} in the previous iteration or a default \textit{truth-value} 
$\left \langle 1,\ iota\right \rangle$ (in the first two iterations), where $iota$ is a hyper-parameter.

\subsection{Matching Module}\label{Matching Module}

In the matching module, first we consider the 1-to-1 range assumption if available (see Appendix \ref{Discussion:1-to-1 Assumption}).

Then the similarity sentences are rearranged.
\textit{Type \uppercase\expandafter{\romannumeral1} path}'s similarities (\textit{type \uppercase\expandafter{\romannumeral1} path}) are naturally sparse, because it only considers the entity pairs which is effectively linked by the logical path. 
Entity name/description's similarities (\textit{type \uppercase\expandafter{\romannumeral2} path}) are dense, however, it is noisy and most of the similarities are useless. 
NALA's similarity inference module exhaustively search for and aggregates the two types of similarity sentences for a specific to-be-aligned entity versus any entity in the other KG.
Then, because of the sparsity of informative similarity signal, the similarity sentences is rearranged into ordered linked list, one list for a specific to-be-aligned entity. 
The sentences are ordered (descending) by its $expectation$ value.
We only store the top $K_{sim}$ similarity sentences in the linked list, where $K_{sim}$ is a hyper-parameter. 

We tackle the EA matching problem as a stable matching problem.
So we propose a recursive bidirectional matching algorithm (rBMat) which has similar idea with BMat ~\cite{RN222}. 
See Algorithm \ref{recursive bidirectional matching} and Algorithm \ref{MatchAndDelete} for details.
The main idea is to recursively find the stable matching of an entity $e_1$ by the "match and delete" function, as well as delete the similarity sentences that don't conform the 1-to-1 assumption.
Considering sorting cost, our rBMat has $O(kn^2)$ time complexity
and $O(kn)$ space complexity, where $k$ represents $K_{sim}$ and $k \ll n$.
For comparison, both BMat and Jonker-Volgenant algorithm have $O(n^3)$ time complexity.

We found that there are still some mismatches after performing rBMat algorithm and most of them share a same pattern. 
For example, $e_{1a} \leftrightarrow e_{2a}$ and $e_{1b} \leftrightarrow e_{2b}$ are two ground truth pairs, however, the result of rBMat is $e_{1a} \leftrightarrow e_{2b}$ and $e_{1b} \leftrightarrow e_{2a}$.
We implement a simple swapping technique to handle this. 
For every pair of similarity sentences, we swap their alignment if the swapped similarity sentences have a higher total $expectation$ value than their original form.
 
\subsection{Unsupervised Learning} \label{Unsupervised Learning}
The seed alignment set is not always available for different EA tasks or real-world EA applications.
So an unsupervised scenario is sometimes adopted to evaluate the industrial applicability of EA methods.
We adapt our method to the unsupervised scenario, that is, without using seed alignments. The BERT embedding model need to finetune on seed alignments, so we adopt a bootstrapping strategy. First, a NALA instance performs alignment on the dataset with 0\% seed and no literal embedding information. 
Then, filter the initial alignment results with an $expectation$ threshold $\theta_{filter}$ and use the filtered results as the finetune training set of BERT. Next, another 0\% seed NALA instance performs alignment with the help of BERT's literal embedding information to obtain the final result.
\section{Interpretability of NALA}
Following ~\cite{RN239, RN236}, interpretable ML (machine learning) focuses on designing models that are inherently interpretable, while explainable ML tries to provide post hoc explanations for existing black box models.
NALA is highly interpretable and self-explanatory. It is arguably more interpretable than PARIS for the following two reasons. First, with the introduction of evidence amount (\textit{confidence}) and logical inference rules, NALA processes data with more information and generates a more informative explanation. Second, NALA manages value similarity, name similarity and structural similarity in a unified logical framework, while PARIS doesn't leverage such side information.

NALA is self-explanatory in the sense that it generates a log file of evidences for the alignments so we can inspect the file after an iteration. 
This feature enhances the troubleshooting capacity of us to some extent during the development process of NALA. 
For example, inspecting the faulty alignments in the evidence file inspired many decision choices in this paper. 
The generated evidences are displayed in our GitHub repository. 

Using the neural BERT model does not weaken the interpretability of \textit{type \uppercase\expandafter{\romannumeral1} path} because utilizing literal value similarity does not affect the interpretable inference steps. 
Moreover, as we only keep the attribute value similarities with a score above the threshold, most of these similarities are easily understood and self-exp\-lanatory, except the wrong ones. Our method tolerates faulty attribute value similarity because \textit{type \uppercase\expandafter{\romannumeral1} path} needs a conjunction of all premises, while faulty similarities usually can't form a complete premise set.

We discuss NALA's relation with other methods and help explain the mechanism of those methods in Appendix \ref{Relation with other methods}.




\section{Experiments and Results} \label{Experiments and Results}
\begin{table*}[t]
  \centering
  \scalebox{0.8}{
  \begin{tabular}{|c|c|cccc|c|c|c|c|}
    \hline  
    \multirow{2}*{Group} & \multirow{2}*{Model} & \multicolumn{5}{|c|}{Settings} & ZH\_EN & JA\_EN & FR\_EN \\
    \cline{3-10}
    ~ & ~ & Attr. & Name & Trans. & Desc. & Seed & Hits@1 & Hits@1 & Hits@1 \\
    \hline
    \multirow{4}*{1} & JAPE & \Checkmark  & & &  & 30\% & 0.412 & 0.363 & 0.324  \\
    ~ & GCNAlign & \Checkmark  & & &  & 30\% & 0.413 & 0.399 & 0.373  \\
    ~ & PARIS+ & \Checkmark  & & &  & 30\% & 0.904 & 0.874 & 0.928  \\
    ~ & NALA & \Checkmark  & & &  & 30\% & \textbf{0.985} & \textbf{0.972} & \textbf{0.990} \\
    \hline
    \multirow{3}*{2} & PARIS & \Checkmark  & & &  & 0\% & 0.777 & 0.785 & 0.793  \\
    ~ & FGWEA* & \Checkmark  & & &  & 0\% & 0.929 &0.922&0.967  \\
    ~ & NALA & \Checkmark  & & &  & 0\% & \textbf{0.982} & \textbf{0.968} & \textbf{0.987} \\
    \hline 
    \multirow{7}*{3} & RDGCN & & \Checkmark & \Checkmark & & 30\% & 0.708 & 0.767 & 0.886 \\
    ~ & CUEA & & \Checkmark & \Checkmark & & 30\% & 0.921 & 0.946 & 0.956 \\
    ~ & UPL-EA & & \Checkmark & \Checkmark & & 30\% & 0.949 & 0.970 & 0.995 \\
    ~ & SE-UEA & & \Checkmark & \Checkmark & & 0\% & 0.935 & 0.951 & 0.957 \\
    ~ & LightEA & & \Checkmark & \Checkmark & & 0\% & 0.952 & 0.981 & 0.995 \\
    ~ & FGWEA* & & \Checkmark & \Checkmark & & 0\% & \textbf{0.959} & 0.982 & 0.994 \\  
    ~ & NALA & & \Checkmark & \Checkmark &  & 0\% & 0.952 & \textbf{0.985} & \textbf{0.996} \\
    \hline
    \multirow{2}*{4} & BERT-INT & \Checkmark & \Checkmark & & \Checkmark & 30\% & 0.968 & 0.964 & 0.995 \\
    ~ & NALA & \Checkmark & \Checkmark & & \Checkmark & 30\% & \textbf{0.998} & \textbf{0.997} & \textbf{0.999} \\
    \hline
    \multirow{3}*{5} & TEA & \Checkmark  & \Checkmark & &  & 30\% & 0.941 &0.941&0.979  \\
    ~ & FGWEA* & \Checkmark  & \Checkmark & &  & 0\% & 0.976 &0.978&0.997  \\
    ~ & NALA & \Checkmark & \Checkmark & &  & 0\% & \textbf{0.993} & \textbf{0.988} & \textbf{0.998} \\ 
    \hline
  \end{tabular}
  
  }
  \caption{Evaluation results of all compared EA methods on DBP15K in five different setting groups. 
Methods marked with * use the additional information of relation names.
We put a supervised method TEA into group 5 for simplicity.}\label{tab:result DBP15K}
\end{table*}
\subsection{Datasets}
We evaluate our model on two EA datasets: the widely used cross-lingual dataset DBP15K (see ~\cite{RN194} for details) and a monolingual multi-source dataset OpenEA benchmarks (including D-W-15K-V2, etc.) ~\cite{RN167}. 
DBP15K consists of three subsets of cross-lingual KG pairs extracted from DBpedia.
Each sub-dataset of OpenEA benchmark consists of two English KGs.
The statistics of the datasets are listed in \autoref{tab:statistics}.

\subsection{Main Results on DBP15K}
\subsubsection{Settings}
The settings of our main results on DBP15K (\autoref{tab:result DBP15K}) consists of five sub-settings: $Attr., Name,\- Trans., Desc.$ and $Seed$, explained as follows.
$Attr.$ is for utilizing the attribute triples. 
$Name$ is for utilizing the entity name information. 
$Trans.$ is for utilizing translators for entity name. We use the Google translator, which is consistent with many other studies.. 
$Desc.$ is for utilizing the information of entity description. 
$Seed$ is for the percentage of seed alignments, 30\% for the conventional supervised scenario and 0\% for the unsupervised scenario. 

We categorize baselines into five setting groups and run NALA using the settings for each group. 
The five groups cover a vast majority of different method's settings.
Group 1 is the supervised scenario with attribute triples.
Group 2 is the unsupervised scenario with attribute triples.
Group 3 is the supervised or unsupervised scenario with entity name information and translator.
Group 4 is the supervised scenario with entity name and description information, which is the same scenario as BERT-INT.
Group 5 is the unsupervised scenario with attribute triples and entity name information.

Most hyper-parameters of our model remain the same across different datasets and setting groups, except for group 3 which will be discussed later. 
The hyper-parameters are selected manually. We set $iota$ = 0.5, $theta$ = 0.1, $C_{penalty} = 4$
and $end\_iteration$ = 19 (20 iterations in total). 
$K_{sim}$ is set to 80. $\theta_{filter}$ is set to 0.9.
The BERT unit is finetuned for 15 epochs. The dimension of the 
BERT CLS embedding is 768 and the dimension of BERT unit's embedding output is 300.

\subsubsection{Main Results}
We compare NALA with the following methods, most of which are new and well-performing: JAPE ~\cite{RN194}, GCNAlign ~\cite{RN196}, PARIS+ ~\cite{RN146}, PA\-RIS ~\cite{RN216}, FGWEA ~\cite{RN163}, RDGCN ~\cite{RN245} 
,CUEA ~\cite{RN225}, UPL-EA ~\cite{RN234}, SE-UEA ~\cite{RN226}, LightEA ~\cite{RN173}, BERT-INT ~\cite{RN168}, TEA ~\cite{RN248}. 
Their results are fetched from their original papers.

 
The experimental settings and results of NALA and all compared baselines on DBP15K are in \autoref{tab:result DBP15K}. 
As observed, NALA achieves the best performance in term of Hits@1 in all five groups except group 3.
NALA outperforms BERT-INT significantly with identical setting and the same embedding method, 
verifying the effectiveness of our similarity inference combined with the matching algorithm.
NALA outperforms FGWEA in group 2 and 5, indicating that it successfully utilizes the information of attribute triples. 
In group 1, two classic EA model JAPE and GCNAlign are outperformed by the newer approaches (PARIS+ and NALA) by
a significant margin, indicating the effective innovation of the new EA approaches in the recent years.
The performance of NALA in unsupervised group 2 approaches its performance in supervised group 1 with a minor gap, 
indicating that 
our proposed bootstrapping strategy effectively adapts to the unsupervised setting (with the help of attribute information).

As for setting group 3, the attribute information is unavailable and we have to rely on the name and translation information
to bootstrap the alignment process. We adjust the hyper-parameter $K_{sim}$ to 400 and other hyper-parameters are unchanged.
We use two BERT units instead of one to separately embed the original entity names and the translated entity names. 
The BERT units are finetuned separately.
We adapt the bootstrapping strategy in Section \ref{Unsupervised Learning} into three steps.
In each step, we perform alignment with a NALA instance and filter the alignment results as the training set of the 
BERT units of the subsequent step. 
The results of each step of the three datasets are shown in \autoref{fig:fig2}.
As expected, the alignment performance increases with the steps, because the BERT unit obtains better finetuning data every step and thus produces better embeddings for alignment.
The results of each iteration of the second bootstrap step are shown in \autoref{fig:fig3}.
NALA outperforms other methods in setting group 3 on JA\_EN and FR\_EN, including three supervised ones.
However, on ZH\_EN unsupervised FGWEA yields better performance. This is possibly 
due to FGWEA's utilization of additional information of relation names. 
The error accumulation effect of NALA's strategy in group 3 is left for further study.

We also conducted preliminary experiments on full version of DBP15K, see Appendix \ref{Appendix: Experimental results on full version of DBP15K}.

\subsection{Results on OpenEA benchmarks}
The original OpenEA benchmark datasets have no meaningful entity names or description, so our results are based on setting group 1.
As shown in \autoref{Experimental results on OpenEA}, NALA achieves better Hits@1 compared with LightEA on OpenEA benchmark datasets except for D-W-15K-V2 and D-W-100K-V2.
However, if attribute information is not available, its performance is worse yet comparable.
It demonstrate that the effectiveness of structural information processing of NALA is likely to be worse than LightEA.
We will explore more type of paths to tackle this problem.


\subsection{Ablation study}
To validate the effectiveness  of each component in NALA, we compare it with several ablations.
We demonstrate the results in ~\autoref{Ablation study table}, where w/o represents without and 
$E_{value}$ represents attribute value embedding information. 
$all\_revision$ represents replacing \textit{pr\-obabilistic revision} rule with \textit{revision} rule 
and $all\_prob\_revision$ is the opposite.
$1-to-1\_range$ is the 1-to-1 matching range information that is utilized in Section \ref{Matching Module} and  
$swapping$ is a proposed technique in Section \ref{Matching Module}. 

NALA performs the best compared with its variants. 
The \textit{revision} rule can deal with negative evidences of similarity sentences, while 
\textit{pr\-obabilistic revision} rule cannot.
The ablation results together with the main results show that NALA seems to have good monotonicity in Hits@1 performance in the sense that when adding extra information or procedure (component) into the model, the Hits@1 increases monotonically. 
Arguably, this is because introducing two-dimensional \textit{truth-values} in every inference step separates \textit{confidence} from truth degree (\textit{frequency}) in every statement, thus the information of relative reliability level is stored for further usage. 

\section{Conclusion and Future Work} \label{Conclusion and Future Work}
In this paper, we propose an entity alignment method named NALA, 
tackling the EA problem by modeling similarity inference and performing a matching algorithm.
Similarity inference obtains similarity through paths that connect the entities.
NALA leverages three type of paths, exploiting both structural and side information of KGs.
Using the similarities, NALA matches the entities by the proposed rBMat algorithm.
NALA is also successfully adapted to the unsupervised scenario and a scenario without attribute triples.
Compared with up-to-date EA methods, NALA attains competitive result on OpenEA benchmark datasets and various settings of DBP15K, 
indicating that it successfully handles the most effective part of similarity inference.


We also take a step in re-evaluating the design choices of different EA models, by providing some interesting 
insights (explanations) of different methods and competitive results compared with them. %
Hopefully, our approach may broaden the view and deepen the understanding of the EA research community. 
How to combine embedding models with path inference and facilitate its full potential is 
a research question to be further studied.


NAL can express and process many different reasoning patterns and logical structures, so NALA can be extended 
to tackle other challenges in the EA process in future research, 
such as integrating ontological information.

\section*{Limitations}
Roughly speaking, NALA has slightly more hyper-parameters than some other EA methods, which may be a drawback.

NALA costs more time compared with the fastest EA methods (1680 seconds compared with 34.5 seconds by LightEA-I on D-Y-100K-V2), possibly due to the inability to utilize GPU in its logical design, thus being more difficult to be parallelized.

The performance of NALA on a hard setting of the datasets, that is, without both attribute triples and entity name information is moderate. Our approach is not yet optimized for utilizing pure structure information of KGs.

\section*{Acknowledgments} The authors sincerely thank the anonymous reviewers for their valuable comments and suggestions, which improved the paper. The work is supported by the National Natural Science Foundation of China (62276057), and Sponsored by CAAI-MindSpore Open Fund, developed on OpenI Community.

\bibliography{custom}

\appendix
\section{Introduction of NAL} \label{Introduction of NAL}
NAL (Non-Axiomatic Logic) ~\cite{RN219} is a logic designed for the creation of general-purpose AI systems, by formulating the fundamental regularities of human thinking in a general level. It can be used as the logical foundation of a (non-axiomatic) inference system and it has been explored to be utilized in various AI tasks ~\cite{RN277, RN278}.
Traditional inference systems are usually based on model-theoretic semantics, while under the assumption of insufficient knowledge and resources, NAL is a term logic basing on
\textit{experience-grounded semantics} ~\cite{RN227}. The meaning of a \textit{term} in NAL, to the inference system, is determined by its role in the \textit{experience} (which will be explained later), that is, how it has been related to other \textit{terms} in the past.
The \textit{truth-value} of a \textit{statement} in NAL is determined by how it has been supported or refuted by other \textit{statements} in the past. 

In this paper we only utilize a fraction of NAL's syntax and inference capability (for EA). We will now introduce the relevant parts of its syntax. 
A \textit{term} in NAL can either be atomic or compound. An \textit{atomic term} is a word (string) or a \textit{variable term}.
\textit{Independent variable}, such as "$\$x$", represents any unspecified \textit{term}
under a given restriction, and intuitively correspond to the universally quantified variable in first-order predicate logic. 
\textit{Dependent variable}, such as "$\#y$", represents a certain unspecified \textit{term} under a given restriction, and intuitively correspond to the existentially quantified variable. 
A \textit{compound term} consists of \textit{term connector} and \textit{components} (which are themselves \textit{terms}).

A basic \textit{statement} has the form of "\textit{subject copula predicate}", where \textit{subject} and \textit{predicate} are \textit{terms}.
There are multiple types of \textit{copula} and each type has a corresponding \textit{statement} type, including: 
1.\textit{Inheritance} ("$A \rightarrow B$", where $A$ and $B$ are \textit{terms}) which intuitively means "B is a general case of A"; 
2.\textit{Similarity} ("$A \leftrightarrow B$") which intuitively means "A is similar with B"; 
3.\textit{Implication}, which is a \textit{higher-order copula} ("$P \Rightarrow Q$", where $P$ and $Q$ are \textit{statements}), intuitively means "P implies Q" (different from the "material implication",
it requires $P$ to be related to $Q$ in content because NAL is a term logic that uses \textit{syllogistic} inference rules and only derives conclusions that are related in content).
A \textit{sentence} is a \textit{statement} together with its \textit{truth-value}.
An \textit{intensional set} with only one component, for example, "$[red]$" intuitively means "red things". 
\textit{Term connector} "$*$" (\textit{product}) combines multiple \textit{component terms} into an ordered \textit{compound term} such as $\left(*,A,B\right)$, 
which intuitively means "an anonymous relation between A and B". 
Compound terms are usually written in the prefix format, that is the \textit{term connector} is written in the first place.
\textit{Statement connector} "$\wedge$" can be seen as the conjunction operator of propositional logic.

NAL is "non-axiomatic" in the sense that the \textit{truth-value} of a conclusion in the inference system does not indicate how much the
conclusion agrees with the "state of affair" in the world, or with a constant set of assumptions (the axioms), but how much it is supported by the
evidence provided by the past \textit{experience} of the system. 
\textit{Experience} means the inference system's history of interaction with the environment or equivalently the input \textit{sentences}. 
The acquisition of \textit{experience} may involve sensorimotor mechanism and sensation-perception process, which is beyond our scope.
The information source of a \textit{sentence} is characterized as its \textit{evidence}.
The inference rules of NAL coherently pass on the evidential information from the premises to the conclusion, 
so the premises can be seen as the \textbf{\textit{evidence}} of the conclusion.
The input \textit{sentences} can be seen as a synthesis of virtual positive and negative evidences.
Assume the available amount of positive evidence and negative evidence of a statement are written as $w^+$ and $w^-$, respectively, then the total amount of evidence is $w = w^++w^-$. The \textbf{\textit{frequency}} of the statement is $f = w^+/w$, and the \textbf{\textit{confidence}} of the statement is $c = w/(w+k)$, 
where $k$ is a positive constant representing "evidential horizon". We take $k$ = 1 in our implementation. 
\textit{Frequency} intuitively means "the degree of truth" and \textit{confidence} intuitively represents "the total amount of evidences".  
The more evidences that the statement have considered, the higher \textit{confidence} value.
The \textbf{\textit{truth-value}} attached to the statement is the ordered pair $\langle f, c\rangle$ and it is often written right after the statement.
The \textbf{\textit{expectation}} of the \textit{truth-value} is a combined measurement of $f$ and $c$, in this paper we define it as $expectation = f \times c$, which is different from its original definition.

NAL uses \textit{syllogistic} (rather than truth-functional) inference rules, that is, the two premises have to share at least one common \textit{term}.  
Among them the \textit{revision} rule merges evidences for the same statement collected from different sources together, so it can settle inconsistency among the system's \textit{sentences}. 
It is very useful in our approach.
The relevant rules with corresponding truth functions are all listed in \autoref{tab:truth functions}. Note that the inference rules are not domain-specific. There are three extended boolean operators ~\cite{RN219} in the calculation of 
truth functions: 

$\left \{ \begin{aligned}
and(x_1,...,x_n) &= \prod_{i=1}^n x_i
\\ or(x_1,...,x_n)\- &= 1 - \prod_{i=1}^n (1 - x_i)
\\ not(x) &= 1 - x
\end{aligned}\right.$
, where $x_i \in [0, 1]$.

\begin{table*}[t]
  \scalebox{0.9}{
  \begin{tabular}{ccc}
    \toprule
    Inference rule & Premises & Conclusion\\
    \midrule
    \multirow{2}*{Deduction} & $A \rightarrow B \ \left \langle f_1,c_1 \right \rangle$ & \multirow{2}*{$A \rightarrow C \ \left \langle f = and(f_1, f_2), c = and(f_1, f_2, c_1, c_2) \right \rangle$} \\
    ~ & $B \rightarrow C \ \left \langle f_2,c_2 \right \rangle$ & ~\\
    \hline
    \multirow{2}*{Analogy}  & $A \rightarrow B \ \left \langle f_1,c_1 \right \rangle$ & \multirow{2}*{$C \rightarrow B \ \left \langle f = and(f_1, f_2), c = and(f_2, c_1, c_2) \right \rangle$} \\
    ~ & $A \leftrightarrow C \ \left \langle f_2,c_2 \right \rangle$ & ~\\
    \hline
    Conditional & $(P\wedge Q)\Rightarrow R\left\langle f_1,c_1 \right \rangle$ & \multirow{2}*{$P \Rightarrow R \ \left \langle f = and(f_1, f_2), c = and(f_1, f_2, c_1, c_2) \right \rangle$} \\
    Deduction & $Q \ \left \langle f_2,c_2 \right \rangle$ & ~\\
    \hline
    \multirow{2}*{Induction} & $A \rightarrow B \ \left \langle f_1,c_1 \right \rangle$ & \multirow{2}*{$C \rightarrow B\left \langle w^+ = and(f_2, c_2, f_1, c_1), w^- = and(f_2, c_2, not(f_1), c_1) \right \rangle$} \\
    ~ & $A \rightarrow C \ \left \langle f_2,c_2 \right \rangle$ & ~\\
    \hline
    \multirow{2}*{Revision} & $P \ \left \langle f_1,c_1 \right \rangle$ & \multirow{2}*{$P \ \left \langle w^+ = w^+_1 + w^+_2, w = w_1 + w_2 \right \rangle$} \\
    ~ & $P \ \left \langle f_2,c_2 \right \rangle$ & ~\\
    \hline
    Probabilistic & $P \ \left \langle f_1,c_1 \right \rangle$ & \multirow{2}*{$P \ \left \langle  f = or(f_1, f_2), w = w_1 + w_2 \right \rangle$} \\
    Revision & $P \ \left \langle f_2,c_2 \right \rangle$ & ~\\
    \bottomrule
  \end{tabular}
  }
  \caption{The table of relevant rules and their truth functions.}\label{tab:truth functions}
\end{table*}

\section{Related Work} \label{Related Works appendix}
\subsection{Embedding-based EA} \label{appendix Embedding-based}
Embedding-based EA methods usually consist of three parts: the embedding module, the alignment module and the matching module.
For the embedding module, translational methods and graph neural network (GNN) methods are the most popular. 
Translational methods, such as MTransE ~\cite{RN195}, usually optimize a margin-based loss function to learn the structural information (relation triples) of a KG.
On the other hand, GNN methods recursively aggregate the representations 
of neighboring nodes with graph convolutional networks (GCNs) or graph attention networks (GATs).
The representative ones are RDGCN ~\cite{RN245} and RREA ~\cite{RN246}, respectively.
The alignment module maps the entity embeddings in different KGs into a unified space. 
There are generally three techniques ~\cite{RN221} for this module: 1. Sharing the embedding space by using the margin-based loss to enforce the
seed alignment entities' embeddings from different KGs to be close. 2. Swapping the triples of seed alignment entities.
3. Mapping the entity vectors from one embedding space to the other using a transformation matrix.
The matching module generates the final alignment result. Common practices use the cosine similarity, the Manhattan distance, or
the Euclidean distance between entity embeddings to measure their similarities and then perform a specific matching
algorithm based on the similarity scores. 
\subsection{Path-based EA} \label{appendix Path-based}
\textbf{PARIS} ~\cite{RN216} is a classic unsupervised non-neural EA method with competitive performance on benchmark datasets ~\cite{RN146}. 
It is purely path-based. 
Following previous works ~\cite{RN276}, PARIS introduces the probabilistic usage of "functionality"
into the field of EA to enhance the validity of similarity inference paths. 
Functionality generally corresponds to the uniqueness of related things, 
for example a man can only have one father but multiple friends, 
so $fun(father)$ is close to 1 and $fun(friend)$ is relatively lower, where $fun()$ represents functionality of a relation or attribute.
See ~\cite{RN216} for more details about functionality.
With functionality, PARIS constructs a probabilistic model that estimates the probabilities
of entity equivalences:

\noindent$Pr\left( x_1 \equiv x_2 \right) = 1 - \prod_{r_1(x_1,y_1),r_2(x_2,y_2)}( 1 - Pr\left(r_1 \subset r_2 \right)\times fun(r_2^{-1})\times$ 
$Pr\left( y_1 \equiv y_2 \right) )$

\noindent As depicted in the above formula, PARIS estimates the equivalence probabilities by
integrating paths that connects corresponding entities.
It also find subrelations between the two ontologies of KG. 
Subrelations, such as $r_1 \subset r_2$, intuitively means a correspondence of two relations of different KGs
such that one relational fact of $r_1$ in $\mathcal{KG}_1$ implies the existence of a corresponding relational fact of $r_2$ in $\mathcal{KG}_2$.
Here is the formula for $Pr\left(r_1 \subset r_2 \right)$:

$\frac{\Sigma_{r_1(x_1,y_1)} \left( 1 - \prod_{r_2(x_2,y_2)}\left( 1 - Pr\left( x_1 \equiv x_2 \right) \times Pr\left( y_1 \equiv y_2 \right) \right) \right)}
{\Sigma_{r_1(x_1,y_1)} \left( 1 - \prod_{x_2,y_2}\left( 1 - Pr\left( x_1 \equiv x_2 \right) \times Pr\left( y_1 \equiv y_2 \right) \right) \right)}$

\noindent With the help of subrelations' measurement, PARIS generalizes the equation of $Pr\left( x_1 \equiv x_2 \right)$ to the
case where the two ontologies do not share common relations.
Therefore, PARIS recursively aligns the entities and the equivalence
probability of $x_1 \equiv x_2$ depends recursively on other equivalence probabilities. In each iteration, the probabilities are re-calcu\-lated
based on the equivalences and subrelations of the previous iteration. Initial equivalences are computed between attribute literals
based on a certain string distance measurement.

\textbf{PARIS+} ~\cite{RN146} is a  variant of PARIS that makes a simple refinement and works in
the absence of attribute triples. It processes the seed alignment information to generate synthetic attribute triples. 
That is, for every pair of seed alignments ($x_1$, $x_2$), it creates the attribute triples ($x_1$, EA:label, string($x_1$)) 
and ($x_2$, EA:label, string($x_1$)), where EA:label is a synthetic relation. Thus, the reverse of the relation EA:label is designed to be
highly functional in order to let the model match the seed alignments easily. NALA adopts the same refinement as PARIS+.

\subsection{Combined EA} \label{appendix Embedding-path}
\textbf{BERT-INT} ~\cite{RN168}, an embedding-path EA meth\-od, uses the well-known transformer model BERT to embed the entities and literals.
It calculates the cosine similarity of the entity name/description embedding.
Then it proposes an interaction model that compares each pair of neighbors or attributes (which forms a path from the source entity to the target entity)
to obtain the neighbor/attribute similarity score. The name/description similarity vector, neighbor similarity vector and attribute similarity vector
are concatenated and applied to a MLP layer to get the final similarity score.

\textbf{FGWEA} ~\cite{RN163} is a three-step progressive optimization algorithm for EA and it can be classified as an embedding-path EA meth\-od.
First, the entity names and concatenated attribute triples are used for semantic embedding
matching to obtain initial anchors. Then in order to approximate GWD (Gromov-Wasserstein Distance\- ~\cite{RN241}), FGWEA computes cross-KG structural and relational similarities, 
which are then used for iterative multi-view optimal transport alig\-nment. Finally, the Bregman Proximal Gradient algorithm ~\cite{RN242} is employed to refine the GWD's coupling matrix.

\subsection{Other EA Methods} \label{appendix other}
There are also a few works that focus on the interpretability or explanability of EA, such as LightEA ~\cite{RN173} and ExEA ~\cite{RN320}.
\textbf{LightEA} is an interpretable non-neural EA method. It is inspired by a classical graph
algorithm, label propagation ~\cite{RN243}. First, it generates a random orthogonal label for each seed alignment entity pair.
Then, the labels of entities and relations are propagated according to the three views of adjacency tensor. 
Finally, LightEA utilizes sparse sinkhorn iteration to address the assignment problem of alignment results.

The ExEA framework, proposed by ~\cite{RN320}, aims to explain the results of embedding-based EA. 
It generates semantic matching subgraphs as explanation by
matching semantically consistent triples around the two aligned entities. ExEA devises an
alignment dependency graph structure to gain deeper insights into the explanation.

The recent literature of EA is abundant, focusing on many different aspects or procedures of entity alignment apart from the aforementioned ones, such as utilizing attribute triples ~\cite{RN179, RN194}, utilizing literals ~\cite{RN207, RN229} , sample mining ~\cite{RN218, RN83}, reinforcement learning ~\cite{RN158}, matching algorithm ~\cite{RN232, RN222, RN157, RN156, RN228}, iterative strategy ~\cite{RN230, RN119} and unsupervised learning ~\cite{RN233, RN226, RN218, RN171, RN225}. There are also some surveys for EA ~\cite{RN221, RN138, RN167, RN173}. Besides graph structural, attribute and literal information, there are other information forms researched by the EA community, such as temporal, spatial and 
graphical information, however, these topics are beyond the scope of this paper. 

\section{Additional Remarks on Inference Paths} \label{Additional Remarks on Inference Paths}

As for type \textit{\uppercase\expandafter{\romannumeral1} path}, we omit two auxiliary inference steps right before arriving at conclusion (6) which performs \textit{structural transformation} in order to dismount $x_2$ from the \textit{product} of (4) without modifying its \textit{truth-value}. The conditional deduction of (11) degenerates into a case without conjunction in its premises (similar with Modus Ponens) and its \textit{truth function} remains the same.
Note that in the path only one direction of the relational inheritance is considered ($r_1 \rightarrow r_2$) and there exists 
a symmetrical variation of the path that utilizes the other direction ($r_2 \rightarrow r_1$). The conclusions of the two paths are aggregated by \textit{probabilistic revision} rule.

\section{Discussion} \label{Discussion}
\subsection{Problem of Understanding Literal Value}\label{Discussion Literal value}
Literal values in real-world KGs act as entity names, entity descriptions, relation/attribute names or attribute values, carrying enormous information. Literal values include texts (strings), numerical values and dates. 
Deep neural network language models provide an interim solution to the problem of understanding literal values. For example, BERT-INT ~\cite{RN168} utilize BERT to embed names/descriptions and values into vector space, thus use similarities between the feature vectors for alignment. 
Literals' deficiency of its outer semantic structure (triples) contrasts with its abundant internal semantics.
However, symbolic reasoning languages (systems) like NAL currently can't effectively handle the subtle semantics in texts for the following reasons: semantic parsing or understanding requires processing capacity and efficiency of complex logical forms and it also requires automatic learning capacity; 
the lacking of KGs with complex logical forms; 
the lacking of KGs with detailed and comprehensive common sense knowledge. In a certain perspective, the literal values in real-world KGs are not really "literal" but rather under-characterized entities, concepts, triples, common sense knowledge and/or statements with complex logical forms. The real-world KG project may not have enough information or adequate paradigm to deal with them. 
For example, the literal value of attribute triple $($John Lennon, deathPlace, "Manhattan, New York City,
United Stat\-es"@en$)$ referred to entities "Manhattan", "New York" and "United \-States", and its form indicates a specific relation between these places.

\subsection{Understanding of \textit{type \uppercase\expandafter{\romannumeral2} path}}\label{Discussion type 2 path}
\textit{Type \uppercase\expandafter{\romannumeral2} path} seems straightforward, however we can have a deeper understanding of it. 
Language models used for the embedding process of EA are distinct information sources other than the KG itself. 
The deep language model which has the ability of aligning or translating entity names can be seen as 
a generalized alignment model that aligns morphemes, words, entities and concepts. 
The pretraining corpus of it consists of sentences, although the sentences do not possesses explicit structures, 
they can be understood or parsed by the model by transforming them into complex logical forms. 
However, such transformation (if exist) and the logical forms are implicitly expressed in the model parameters and 
intermediate layer vector representations.
To summarize, our similarity inference's \textit{type \uppercase\expandafter{\romannumeral2} path} can be seen as the aggregation of multiple virtual complex logical paths.
The aggregation result is represented into the vector space by the language model.

\subsection{1-to-1 Assumption} \label{Discussion:1-to-1 Assumption} 
There are 1-to-1 assumptions in some EA datasets (such as $DBP\-15K$) and it is a useful information for alignment. Formally, we define the 1-to-1 assumption as follows: 
first, there is a range of alignable entities $A_1\subset\mathcal{E}_1$ and $A_2\subset\mathcal{E}_2$ (for DBP15K, $A_1 \subsetneqq \mathcal{E}_1$). Second, the equivalence between $A_1$ and $A_2$ is a bijection. 
Note that the assumption does not have aligning regularity for entities outside the range except that they can't be aligned with entities inside the range.
Many ranking-based EA methods leverages the 1-to-1 range assumption, however, NALA do not.
Therefore, in implementation in order to leverage the range assumption 
we take the set $A_1$ and $A_2$ as input and filters out any alignment sentence that aligns 
$A_1$ to $\mathcal{E}_2 \setminus A_2$ or $\mathcal{E}_1 \setminus A_1$ to $A_2$.

\subsection{Relation with Other Methods} \label{Relation with other methods} 
In this section, we will discuss the relation between our proposed method and methods with other forms.
We will propose some preliminary explanations of certain translational embedding methods and embedding-path EA methods from a theoretical perspective.

The way NAL models KG information and the inference process has a similar part with "uncertainty estimation" ~\cite{RN217} 
in the natural language processing domain.
The \textit{truth-value} of alignments shares some similarity with the distributive view of facts or beliefs which views facts as probability distribution of random variables. Also, the concept of confidence is shared with some information extraction systems such as Markov logic network ~\cite{RN209}, which assigns confidence to extracted facts or logical formulas in some intermediate steps.

\subsection{Relation with Translational Embedding Methods}
The well-known KG embedding model TransE~\cite{RN192} is initially proposed for link prediction tasks.
It may be partially explained from a logical perspective of NAL (or equivalently other logic with similar expressive power). Consider a specific type of Horn clauses $((*,A,B)\rightarrow R_1 \wedge\ (*,B,C)\rightarrow R_2) \ \Rightarrow (*,A,C)\rightarrow R_3 \ \left \langle f_1,c_1 \right \rangle$, the following three triples
\begin{gather*}
(Martin\_Luther\_King\_Jr,\ birthPlace,\ Georgia\_(U.S.\_state)) \\
(Georgia\_(U.S.\_state),\ country,\ United\_States) \\
(Martin\_Luther\_King\_Jr,\ citizenship,\ United\_States)
\end{gather*}
together forms a piece of positive evidence of an instantiated Horn clause, in which $R_1$, $R_2$ and $R_3$ is replaced by $birthPlace$, $country$ and $citizenship$ respectively. We conjecture that the gradient descent optimization process of TransE implicitly performs approximate logical inference and evidence aggregation. In the above example for each of the three triples, $|| \textbf{h} + \textbf{r} - \textbf{t} ||$ (where bold format represent a vector) is minimized once per epoch (ignoring margin-based criterion), leading to $\textbf{birthPlace} + \textbf{country} \approx \textbf{citizenship}$. Thus, the instantiated Horn clause together with its \textit{truth-value} may be represented by the correlation of vector representations, and the \textit{truth-value} may be reflected in distance $||\textbf{birthPlace} + \textbf{country} - \textbf{citizenship}||$. Note that these three relations may appear in more than one Horn clauses, so the gradients from the evidences of a Horn clause may confuse with (or conflict with) those from another Horn clause, for example $\textbf{manufacturer} + \textbf{country} \approx \textbf{made-} \\\textbf{InCountry}$. The training process may force vector $\textbf{birthPlace}$ to be dissimilar with $\textbf{manufacturer}$, otherwise, there may be hallucination in link prediction or EA results. A similar explanation of hallucination may apply to LLMs. A similar analysis applies to the vector representations of two relations which frequently appear on the same head entity (or tail entity).
It's arguable that the test set link prediction process of TransE mainly relies on Horn clauses, because from a logical perspective there is scarcely any other information. In this paper Horn clauses will not be extracted and managed, leaving for further research.

MTransE ~\cite{RN195} is a translational embedding-based EA method. It encodes the two KGs' relational triples separately with 
the TransE loss criterion $S_K = \Sigma_{(h,r,t)}|| \textbf{h} + \textbf{r} - \textbf{t} ||$. It proposed a "distance-based axis calibration"
alignment model in order to coincide the vectors of counterpart entities or relations. The corresponding loss is 
$S_{a_2} = \Sigma|| \textbf{e}_1 - \textbf{e}_2 || + || \textbf{r}_1 - \textbf{r}_2 ||$ ($S_{a_2}$ only has the first item if there is no available 
seed relation alignment). The seed and derived alignments are assumed to have $\textbf{e}_1 \approx \textbf{e}_2$ and we see it as the 
embedding representation of the similarity statement $e_1 \leftrightarrow e_2$, with its \textit{truth-value} somehow represented by the 
distance $|| \textbf{e}_1 - \textbf{e}_2 ||$. Theoretically, the distance can't simultaneously represent \textit{frequency} and \textit{confidence} 
by itself, but more possibly a combined effect.
We argue that MTransE performs approximate inference 
that is similar with the \textit{type \uppercase\expandafter{\romannumeral3} path}, because if the learned embedding 
constraints of the four premises are considered simultaneously, we can get $\textbf{r}_1 \approx \textbf{r}_2$ which 
we interpret as $r_1 \leftrightarrow r_2$. Similarly, MTransE performs approximate inference 
of the \textit{type \uppercase\expandafter{\romannumeral1} path} (with functionality omitted and 
$r_1 \rightarrow r_2$ replaced by $r_1 \leftrightarrow r_2$) to obtain derived alignment results.

\subsection{Relation with Embedding-path EA Methods} \label{Relation with Embedding-path EA Methods}
Here we propose some preliminary explanations of the similarity inference aspect of some embedding-path EA methods from a theoretical perspective.

The first method to be discussed is BERT-INT. It generates entity embedding using the name/description information with BERT unit and the embedding is $C(e) = MLP(CLS(e))$.
It uses pairwise margin loss to approximately enforce $C(e) \approx C(e')$.
Different from MTransE which performs path inference implicitly with the gradient optimization of loss criterions, 
BERT-INT explicitly performs path inference with its proposed interaction model.
Every element of the neighbor-view interaction matrix represents a inference process of a \textit{type \uppercase\expandafter{\romannumeral1} path}.
Its path omits functionality and relation alignment (for BERT-INT fails to utilize its proposed relation mask matrix).
Because of the ignorance of relation type, its premise (1) and (4) has the form of $\left(*,x_1,y_1\right) \rightarrow \#r$ and 
$\left(*,x_2,y_2\right) \rightarrow \#r$ which represents "There exists an unspecified relation between $x_1$/$y_1$, and (another) 
unspecified relation between $x_2$/$y_2$". 
Moreover, its premise (5) fails to utilize derived alignments, because BERT-INT is not iterative.
With such premises, BERT-INT's \textit{type \uppercase\expandafter{\romannumeral1} path} inference's 
effectiveness is supposed to be lower than that of NALA's.
Similarly, every element of the attribute-view interaction matrix represents a \textit{type \uppercase\expandafter{\romannumeral1} path} which has attribute 
triples as premises (1) and (4).
BERT-INT's evidence aggregation method is different from NALA which uses \textit{probabilistic revision} and \textit{revision} rules.

The second method to be discussed is FGWEA. Its multi-view Optimal Transport (OT) alignment step combines
four cost matrices for the OT problem, that is, $C_{sum} = C_{stru} + C_{rel} + C_{name} + C_{attr}$.
Obtaining the cost matrices corresponds to the similarity inference process and different matrices 
correspond to different groups of inference paths.
Among them, $C_{rel}$ corresponds to a degenerated \textit{type \uppercase\expandafter{\romannumeral1} path} inference where 
relation alignment is obtained by relation names and without the consideration of functionality.
$C_{stru}$ corresponds to a further degenerated \textit{type \uppercase\expandafter{\romannumeral1} path} inference 
(similar with BERT-INT's neighbor-view interaction).
$C_{name}$ corresponds to \textit{type \uppercase\expandafter{\romannumeral2} path} inference. 
$C_{attr}$ fails to model the (fine-grained) attributive \textit{type \uppercase\expandafter{\romannumeral1} path}
because it uses the concatenation of all attribute triples of an entity.

In this paper, BERT-INT and FGWEA are classified as embedding-path EA methods because
their embedding module couples with the path inference to some extent. 
In contrast, NALA, which we classify as path-based, performs path inference wherever it can and uses embeddings minimally.

\section{Experiments} \label{appendix Experiments}
\begin{table}[hb]
  
  \scalebox{0.75}{
  \begin{tabular}{ccccc}
    \toprule
    Dataset & $\vert\mathcal{E}\vert$ & $\vert\mathcal{R}\vert$ & $\vert\mathcal{T_R}\vert$ & $\vert\mathcal{T_A}\vert$\\
    \midrule
    \multirow{2}*{$DBP15K_{ZH\_EN}$} & 19,388 & 1,701 & 70,414 & 379,684\\
    ~ & 19,572 & 1,323 &  95,142 & 567,755\\
    \multirow{2}*{$DBP15K_{JA\_EN}$} & 19,814 & 1,299 & 77,214 & 354,619\\
    ~ & 19,780 & 1,153 &  93,484 & 497,230\\
    \multirow{2}*{$DBP15K_{FR\_EN}$} & 19,661 & 903 & 105,998 & 528,665\\
    ~ & 19,993 & 1,208 &  115,722 & 576,543\\
    \hline
    \multirow{2}*{$DBP15Kfull_{ZH\_EN}$} & 66,469 & 2,830 & 153,929 & 379,684 \\
    ~ & 98,125 & 2,317 & 237,674 & 567,755 \\
    \multirow{2}*{$DBP15Kfull_{JA\_EN}$} & 65,744 & 2,043 & 164,373 & 354,619 \\
    ~ & 95,680 & 2,096 &  233,319 & 497,230 \\
  \bottomrule
\end{tabular}
}
\caption{Dataset statistics. $\vert\mathcal{E}\vert$, $\vert\mathcal{R}\vert$, $\vert\mathcal{T_R}\vert$ and $\vert\mathcal{T_A}\vert$ represent the number of entities, relation types, relation
triples and attribute triples in each KG, respectively. 
The statistics of OpenEA benchmark datasets are in ~\cite{RN167}}\label{tab:statistics}
\end{table}

\subsection{Evaluation Metric \& Environment }
We use Hits@1 (which is the same metric as recall for EA) as the sole evaluation metric of our main results of DBP15K for the following reasons. Mean Reciprocal Rank (MRR) is unavailable for NALA because it does not provide a alignment ranking for the test entities. There exist a non-negligible number of equivalent entity pairs that are not in the ground-truth set of DBP15K, so the precision and F1-score can't be measured properly. 
We use the precision (P), recall (R), and F1 score in the ablation study of OpenEA benchmark datasets.

Our NALA model is implemented in java and the BERT unit is implemented in python with PyTorch. All experiments are performed on a
Linux server with an Intel(R) Xeon(R) Silver 4210R CPU @ 2.40GHz, 251G RAM and a NVIDIA GeForce RTX 3090 GPU.

\subsection{Experimental Results on Full Version of DBP15K} \label{Appendix: Experimental results on full version of DBP15K}

The aforementioned and widely used DBP15K dataset version is reduced from a full version. 
That is, any triple that does not contain an entity in the 15k-15k 1-to-1 range is discarded, 
resulting in ~19k entities in both side of KG.
Thus, the full version has more entities and relation triples.
Most embedding-based EA methods rely on the 1-to-1 range information to generate the rankings for alignment.
For the full version of DBP15K, such information may have a bigger effect on EA performance.
As shown in \autoref{Experimental results on full version of DBP15K}, to our knowledge only few methods have conducted
experiments on the full version, yet none of them is implemented without 1-to-1 range. The results of other methods are from ~\cite{RN240} (RPR-RHGT). 
NALA has reasonable Hits@1 performance even without the help of 1-to-1 range information and greatly surpasses other listed methods.

\begin{table}[h]
  \scalebox{0.8}{
  \begin{tabular}{c|c|c}
    \toprule
    Method & $full_{ZH\_EN}$ & $full_{JA\_EN}$ \\
    \hline
    NALA(without 1-to-1 range) &     \textbf{0.908}   &   \textbf{0.889}    \\
    \hline
    JAPE(with 1-to-1 range) &    0.264   &   0.238     \\
    RDGCN(with 1-to-1 range) &    0.621   &   0.812     \\
    RPR-RHGT(with 1-to-1 range) &    0.693   &    0.886   \\
    NALA(with 1-to-1 range) &    \textbf{0.966}   &    \textbf{0.946}   \\
  \bottomrule
\end{tabular}
}
\caption{Experimental results (Hits@1) on full version of DBP15K. The experiments for NALA are based on setting group 1, without attribute value embeddings.}\label{Experimental results on full version of DBP15K}
\end{table}

\subsection{Influence of Confidence Hyper-parameter}

\begin{figure}
  \centering
  \includegraphics[width=\linewidth]{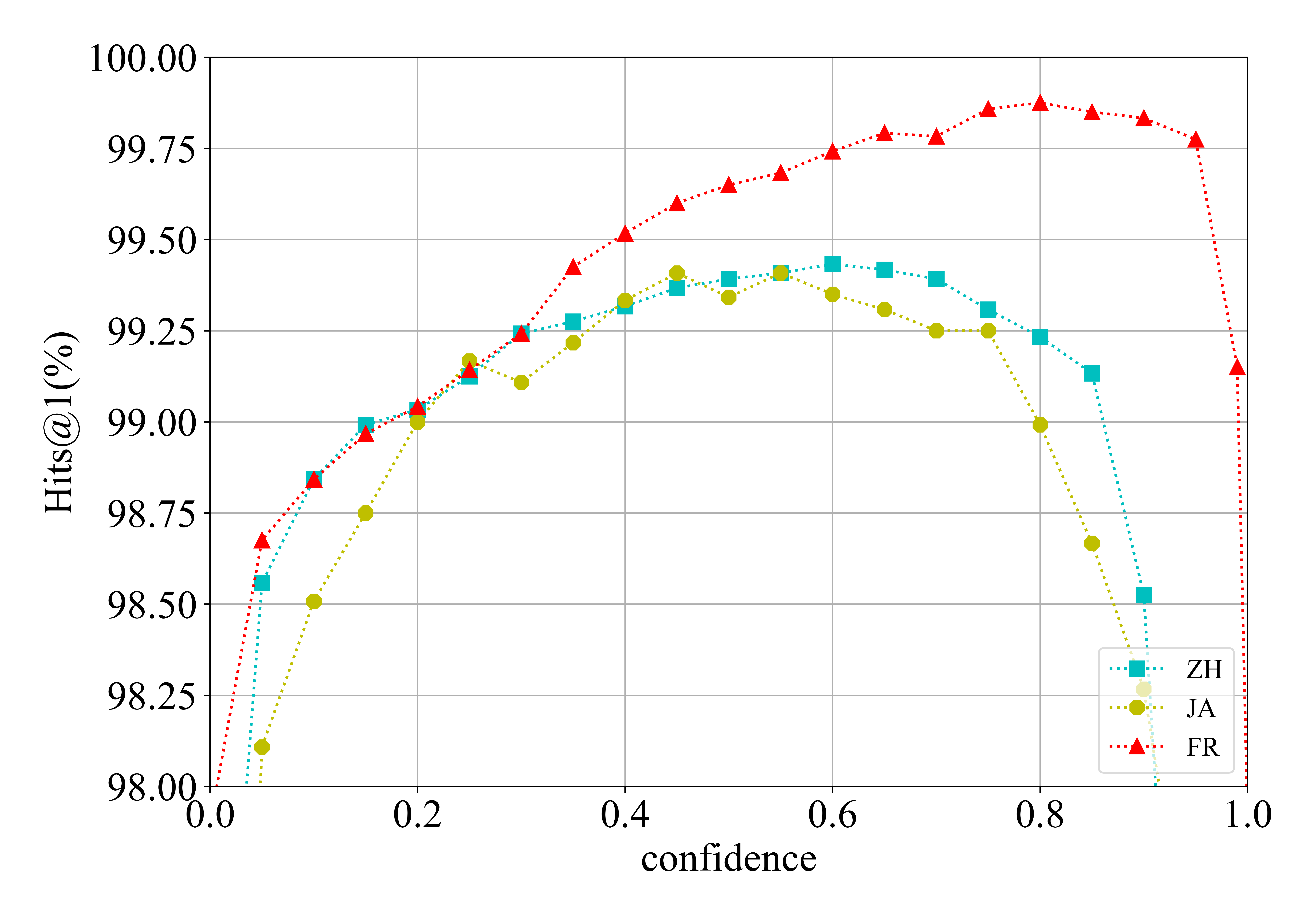}
  \caption{Influence of $C_{name}$.}
  \label{fig:fig_emb_confidence}
\end{figure}

The experiment results of \autoref{fig:fig_emb_confidence} shows how entity name/descrip\-tion embedding similarity \textit{confidence} $C_{name}$ 
(without the adaptive setting of $C_{name}$) affects Hits@1. 
These experiments are performed on setting group 4 without using attribute value embedding information.
We adjust $C_{name}$ with other conditions unchanged. The Hits@1 curve is approximately concave and for 
$ZH\_EN$, $JA\_EN$ and $FR\_EN$ respectively, it reaches maximum performance at 0.6, 0.55 and 0.8. It shows that the informative embedding similarity enhances the performance to different extents. 
French is often regarded as more closely related to English than Chinese or Japanese, so the BERT unit learns representation easier and thus produces more confident embedding similarity.
Pretraining corpus of the BERT unit may include relevant triples (in the form of natural language sentences) which may have same informational origin with DBpedia. So the embedding similarity's evidences may have an overlap part with \textit{type \uppercase\expandafter{\romannumeral1} path}'s evidences.
The \textit{revision} rule is only appropriately used when the two premises don't share same evidence (or equivalently their evidential bases do not overlap). 
So the appropriate \textit{confidence} value need to be lower than the \textit{confidence} of the BERT output (if it provides such information) in order to exclude the overlap. The best-performance \textit{confidence} of each dataset is conjectured to reflect the combined influence of embedding quality of the BERT unit and the evidence overlapping effect. 
The $C_{name}$ \textit{confidence} value can be alternatively set equal to the cosine similarity of the embeddings, 
resulting in a slightly decreased performance. 
This is another good choice if you want to avoid hyper-parameter tuning.

\begin{table}[h] 
  \scalebox{0.8}{
  \begin{tabular}{c|c|c|c}
    \toprule
    Dataset & LightEA-I & NALA & NALA(w/o attr)  \\
    \hline
    D-W-15K-V1 &    0.732   &   \textbf{0.780}    &  0.625  \\
    D-W-100K-V1 &    0.642   &    \textbf{0.732}   &  0.528  \\
    D-Y-15K-V1 &    0.826   &    \textbf{0.949}   &   0.751  \\
    D-Y-100K-V1 &     0.781   &   \textbf{0.927}    &   0.700  \\
    \hline
    D-W-15K-V2 &    \textbf{0.951}   &   0.938    &   0.902  \\
    D-W-100K-V2 &   \textbf{0.926}    &    0.919   &  0.857  \\
    D-Y-15K-V2 &    0.976   &   \textbf{0.983}    &   0.975  \\
    D-Y-100K-V2 &   0.977    &    \textbf{0.984}   &  0.913  \\
  \bottomrule
\end{tabular}
}
\caption{Experimental results (Hits@1) on OpenEA benchmarks. The experiments for NALA are done without attribute value embeddings.}\label{Experimental results on OpenEA}
\end{table}

\begin{figure}[H]
  \centering
  \includegraphics[width=\linewidth]{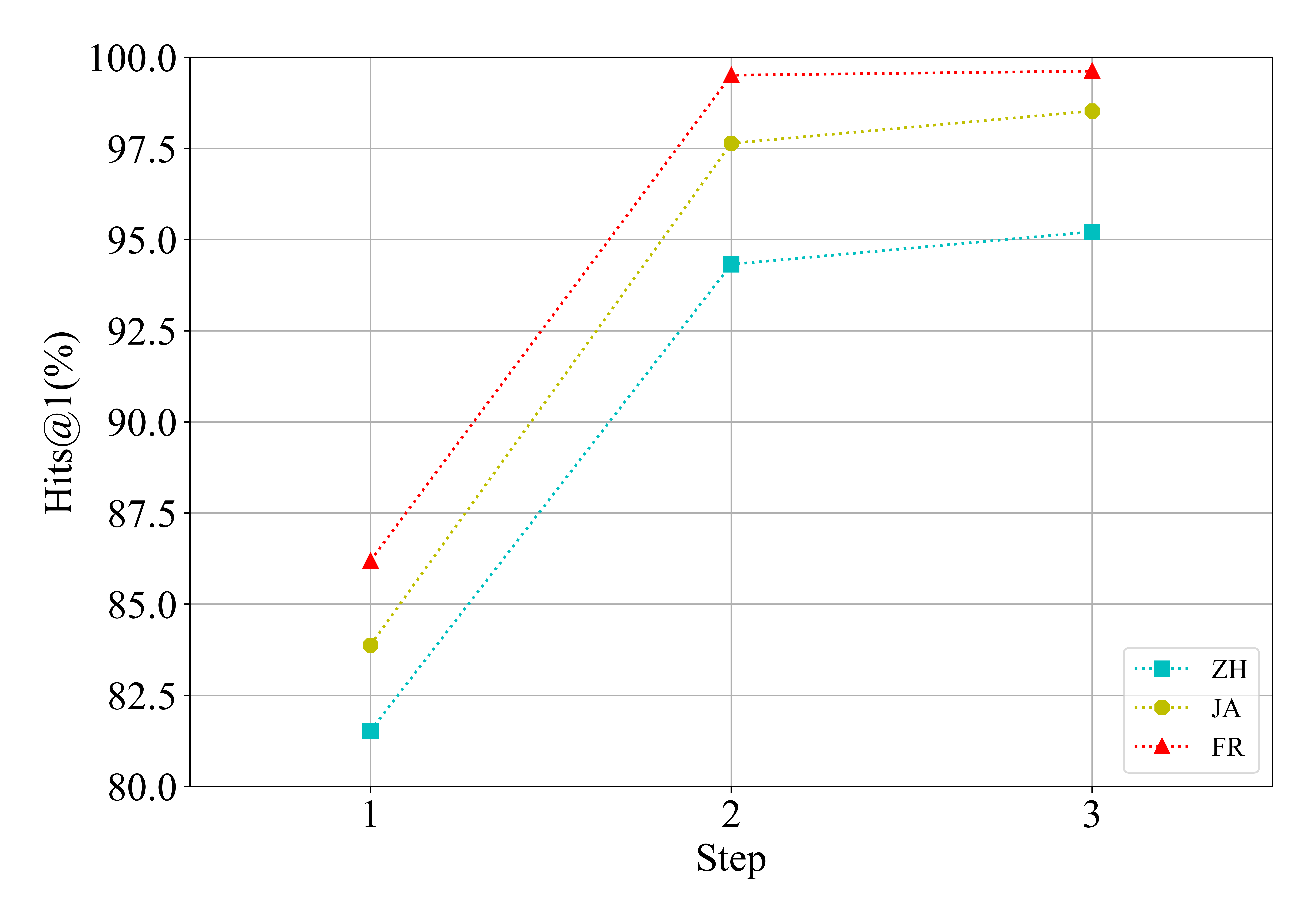}
  \caption{Results of bootstrap steps of setting group 3.}
  \label{fig:fig2}
\end{figure}

\begin{figure}[H]
  \centering
  \includegraphics[width=\linewidth]{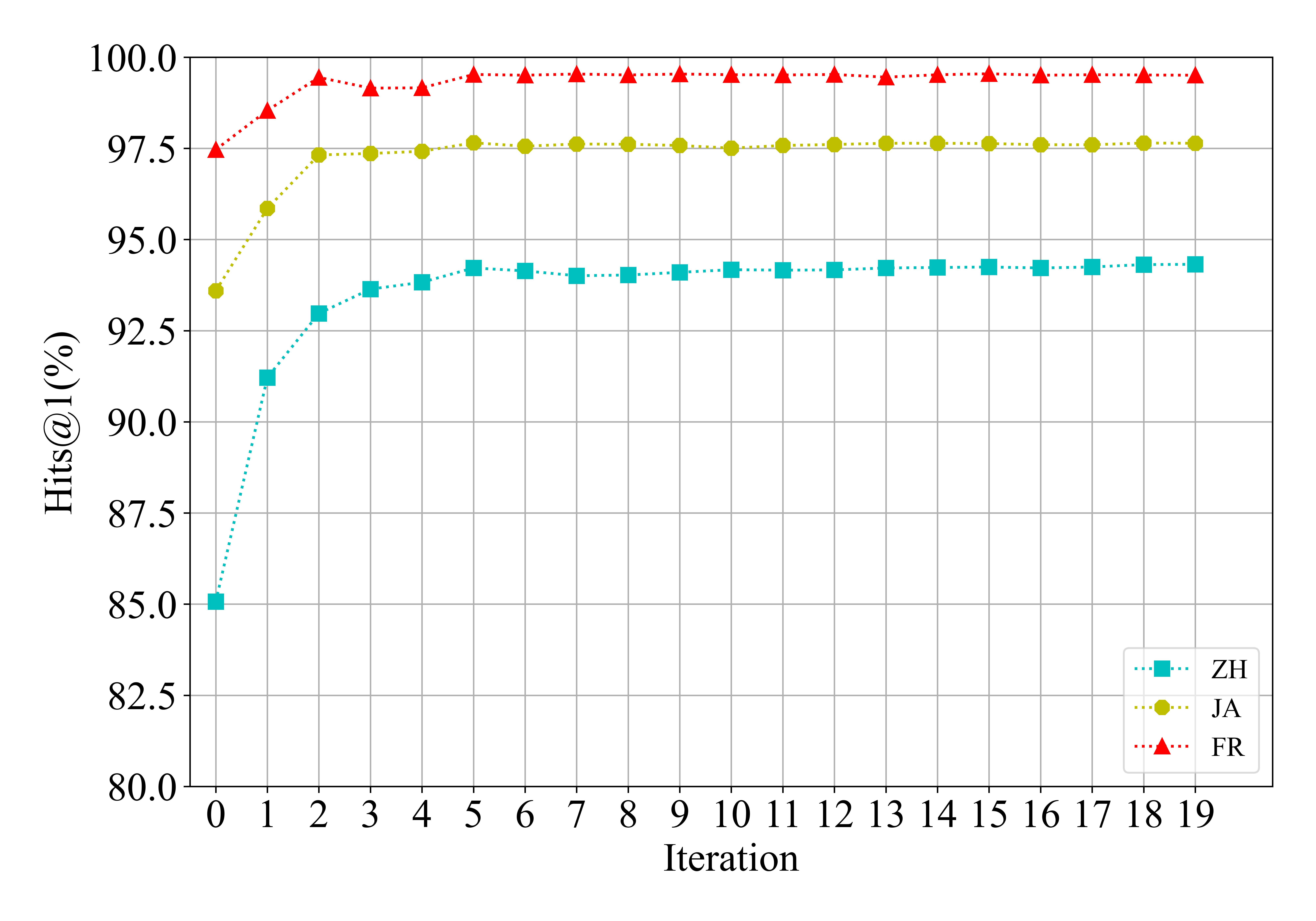}
  \caption{Results of the iterations of the second bootstrap step in setting group 3.}
  \label{fig:fig3}
\end{figure}

\section{Algorithms} \label{Algorithms}
\IncMargin{1em}
\begin{algorithm}
\small
\caption{recursive bidirectional matching} \label{recursive bidirectional matching}  
      \SetKwData{Left}{left}\SetKwData{This}{this}\SetKwData{Up}{up} \SetKwFunction{Union}{Union}\SetKwFunction{MatchAndDelete}{match\_and\_delete} \SetKwInOut{Input}{input}\SetKwInOut{Output}{output}
	
	\Input{An array of linked list of \textit{similarity sentences} $KG1\_to\_KG2$, 
           with each linked list storing top-k \textit{similarity sentences} of an entity with descending order.} 
	\Output{Optimized 1-to-1 \textit{similarity sentences} (alignment results)}
	 \BlankLine 
      \emph{populates $KG2\_to\_KG1$ with all of the sentences in $KG1\_to\_KG2$}\;
      \tcc{$KG2\_to\_KG1$ is another array of linked list, arranging the similarity sentences in the other direction}
	 \For{$e_1$ \textbf{in} $\mathcal{E}_1$}{ 
            \MatchAndDelete{$e_1, null$}\;
 	 } 
 	 	    
 	 \end{algorithm}

\begin{algorithm} 
\small
\caption{match and delete} \label{MatchAndDelete} 
\SetKwData{Left}{left}\SetKwData{This}{this}\SetKwData{Up}{up} \SetKwFunction{Union}{Union}\SetKwFunction{MatchAndDelete}{match\_and\_delete} \SetKwInOut{Input}{input}\SetKwInOut{Output}{output}
	
	\Input{Entity $e_1$, entity $e_{prev}$.} 
        \tcc{$e_1$ is the entity to be matched and we assume that $e_1$ belongs to $\mathcal{KG}_1$, similarly otherwise. Entity $e_{prev}$ represents the previous entity, that is the concerned entity of the recursion parent.}
	\Output{entity $e_{match}$ which forms a stable matching with $e_1$}
	 \BlankLine 
	 \For{$sentence$ \textbf{in} $KG1\_to\_KG2(e_1)$}{ 
            $e_2$ $\leftarrow$  $predicate\_term$ of $sentence$\;
            \tcc{$predicate\_term$ means the other entity of the similarity sentence}
            \If{$e_2$ == $e_{prev}$}{
                $e_{match}$ $\leftarrow$ $e_{prev}$\;
                \textbf{break}\;
            }
            \Else{
                $e_3$ $\leftarrow$ \MatchAndDelete{$e_2, e_1$}\;
                \If{$e_3$ == $e_1$}{
                    $e_{match}$ $\leftarrow$ $e_2$\;
                    \textbf{break}\;
                }
            }
 	 } 
       \For{$sentence$ \textbf{in} $KG1\_to\_KG2(e_1)$ except the first node}{
           \tcc{now that the first sentence for $e_1$ is bidirectionally matched, we delete other sentences}
           \emph{removes $sentence$ from the linked list}\;
           \emph{removes $sentence$'s counterpart in $KG2\_to\_KG1$ which expresses the same similarity in the other direction}\;
       }
       \KwRet{$e_{match}$}\;
 	 	  
 	 \end{algorithm}
 \DecMargin{1em} 

\IncMargin{1em}
\begin{algorithm} 
\small
\caption{NALA(supervised)} \label{NALA(supervised)}
\SetKwData{Left}{left}\SetKwData{This}{this}\SetKwData{Up}{up} \SetKwFunction{Union}{Union}\SetKwFunction{RecursivelyDelete}{recursively\_delete} \SetKwInOut{Input}{input}\SetKwInOut{Output}{output}
	
	\Input{Two knowledge graphs $\mathcal{KG}_1$ and $\mathcal{KG}_2$.} 
	\Output{Alignment result and other information.}
	 \BlankLine 
      \emph{run finetuning for BERT unit}\;
      \emph{compute entity/value embeddings with the BERT unit}\;
      \emph{generate synthetic attribute triples for seed alignments (for supervision)}\;
      \emph{load the knowledge graphs}\;
      \For{$iteration\leftarrow 0$ \KwTo $end\_iteration$}{ 
            \For{$y_1$ \textbf{in} $\mathcal{E}_1$}{ 
                \tcc{aligning for different entities of $\mathcal{E}_1$ is divided into multiple parallel threads}
                \For{$x_1$, $x_2$, $y_2$ that forms a sound \textit{type \uppercase\expandafter{\romannumeral1} path} with $y_1$ (depth-first)}{ 
                    \emph{perform inference of \textit{type \uppercase\expandafter{\romannumeral1} path}}\;
                    \emph{perform inference of \textit{type \uppercase\expandafter{\romannumeral3} path}}\;
         	 } 
                \For{$y_2$ \textbf{in} $\mathcal{E}_2$}{ 
                    \emph{retrieve embedding similarity for $y_1 \leftrightarrow y_2$}\;
                    \emph{perform inference of \textit{type \uppercase\expandafter{\romannumeral2} path}}\;
                }
                \emph{filter the similarity sentences with 1-to-1 range assumption}\;
                \emph{insert the sentences into a top-k ordered linked list}\;
     	 }
            \emph{perform recursive bidirectional matching}\;
            \emph{swapping}\;
            \emph{save alignment results and evidence log file}\;
 	 }
 	 \end{algorithm}
 \DecMargin{1em}

\begin{table*}[hb] 
  \scalebox{0.8}{
  \begin{tabular}{c|c|c|ccc|ccc|ccc}
    \toprule
    \multirow{2}*{Model} & ZH\_EN & JA\_EN & \multicolumn{3}{c|}{D-W-15K-V2} & \multicolumn{3}{c|}{D-Y-15K-V2} & \multicolumn{3}{c}{D-Y-100K-V2}\\
    ~ & Hits@1 & Hits@1 & P & R & $F_1$ & P & R & $F_1$ & P & R & $F_1$ \\
    \midrule
    NALA             & \textbf{0.993} & \textbf{0.988} & 0.917 & \textbf{0.908} & 0.912 & \textbf{0.983} & \textbf{0.981} & \textbf{0.982} & \textbf{0.985} & \textbf{0.980} & \textbf{0.983} \\
    - w/o $E_{value}$ & 0.980 & 0.980 & - & -  & - & - & -  & - & - & -  & - \\
    - $all\_revision$ & 0.964         & 0.912          & 0.857 & 0.814          & 0.835 & 0.899     & 0.871     & 0.885  & 0.402 & 0.312  & 0.351 \\
    - $all\_prob\_revision$ & 0.985 & 0.987 & - & -  & - & - & -  & - & - & -  & - \\
    - w/o $1-to-1\_range$ & 0.989 & 0.978 & - & -  & - & - & -  & - & - & -  & - \\
    - w/o $swapping$ & 0.991 & 0.982 & 0.912 & 0.901  & 0.907 & 0.975 & 0.972  & 0.973  & 0.981 & 0.976  & 0.978 \\ 
    \hline
    FGWEA & 0.976 & 0.978 & \textbf{0.952} & 0.903 & \textbf{0.927}  & - & -  & - & - & -  & - \\
  \bottomrule
\end{tabular}
}
\caption{Ablation study of NALA. To be consistent with FGWEA, the experiments for NALA on DBP15K are based on setting group 5.
The experiments for NALA on OpenEA benchmarks are based on setting group 2, without attribute value embeddings.}\label{Ablation study table}
\end{table*}

\end{CJK*}
\end{document}